\pdfoutput=1

\documentclass[11pt]{article}

\usepackage[]{acl}

\usepackage{times}
\usepackage{latexsym}

\usepackage[T1]{fontenc}

\usepackage[utf8]{inputenc}

\usepackage{microtype}

\usepackage{inconsolata}

\usepackage[final]{pdfpages}

\usepackage{float}

\usepackage{array}
\newcolumntype{P}[1]{>{\centering\arraybackslash}p{#1}}
\usepackage{colortbl}
\usepackage{arydshln}
\usepackage{times,latexsym}
\usepackage{url}
\usepackage[T1]{fontenc}
\usepackage[utf8]{inputenc} 
\usepackage{nicematrix}
\usepackage{booktabs}  
\usepackage{multirow}  
\usepackage{mathtools}  
\usepackage{amsmath}  
\usepackage{enumitem}
\usepackage{nicefrac}

\usepackage[normalem]{ulem}

\def\ODdel#1{\bgroup\markoverwith{\textcolor{purple!60}{\rule[0.4ex]{2pt}{3pt}}}\ULon{#1}}

\def\OPdel#1{\bgroup\markoverwith{\textcolor{orange!60}{\rule[0.4ex]{2pt}{3pt}}}\ULon{#1}}

\def\ZKdel#1{\bgroup\markoverwith{\textcolor{magenta}{\rule[0.4ex]{2pt}{3pt}}}\ULon{#1}}

\def\KOdel#1{\bgroup\markoverwith{\textcolor{blue!60}{\rule[0.4ex]{2pt}{3pt}}}\ULon{#1}}

\title{FreshTab: Sourcing Fresh Data for Table-to-Text Generation Evaluation}

\author{
Kristýna Onderková, Ondřej Plátek, Zdeněk Kasner \and Ondřej Dušek \\
Charles University, Faculty of Mathematics and Physics \\ Institute of Formal and Applied Linguistics \\ Prague, Czechia \\
\texttt{\{onderkova,oplatek,kasner,odusek\}@ufal.mff.cuni.cz} \\
}

\usepackage{listings}

\begin{document}
\maketitle

\begin{abstract}
Table-to-text generation (insight generation from tables) is a challenging task that requires precision in analyzing the data.
In addition, the evaluation of existing benchmarks is affected by contamination of Large Language Model (LLM) training data as well as domain imbalance.
We introduce \textit{FreshTab}, an on-the-fly table-to-text benchmark generation from Wikipedia, to combat the LLM data contamination problem and enable domain-sensitive evaluation.
While non-English table-to-text datasets are limited, \textit{FreshTab} collects datasets in different languages on demand (we experiment with German, Russian and French in addition to English). 
We find that insights generated by LLMs from recent tables collected by our method appear clearly worse by automatic metrics, but this does not translate into LLM and human evaluations. Domain effects are visible in all evaluations, showing that a~domain-balanced benchmark is more challenging.

\end{abstract}

\section{Introduction}
\label{sec:intro}

Table-to-text generation or insight generation \cite{liu_table--text_2018,parikh-etal-2020-totto} is a challenging task in natural language generation (NLG), where a NLG system generates insights from a data table.
This can provide important support in data analytics and decision making in business or governance.
Recent research in insight generation builds on finetuned neural language models \cite{nan-etal-2022-r2d2,zhao-etal-2023-loft, kantharaj-etal-2022-chart} or prompted large language models (LLMs) \cite{zhao-etal-2023-investigating, bian_helm_2024}.

LLMs display excellent performance in various tasks, and unlike prior methods, they do not require costly in-domain training data with human-written references. 
With few-shot examples and chain-of-thought prompting, they surpass prior methods on insight generation \cite{zhao-etal-2023-investigating}.
However, LLMs were also shown to memorize common benchmarks \cite{oren_proving_2024, xu2024benchmark}, inflating their true performance, and to perform unevenly across domains \cite{hu-etal-2024-large-language,diao_guidebench_2025,zhu_domaineval_2025}.

\begin{figure}[t]
    \centering
    \includegraphics[width=0.45\textwidth]{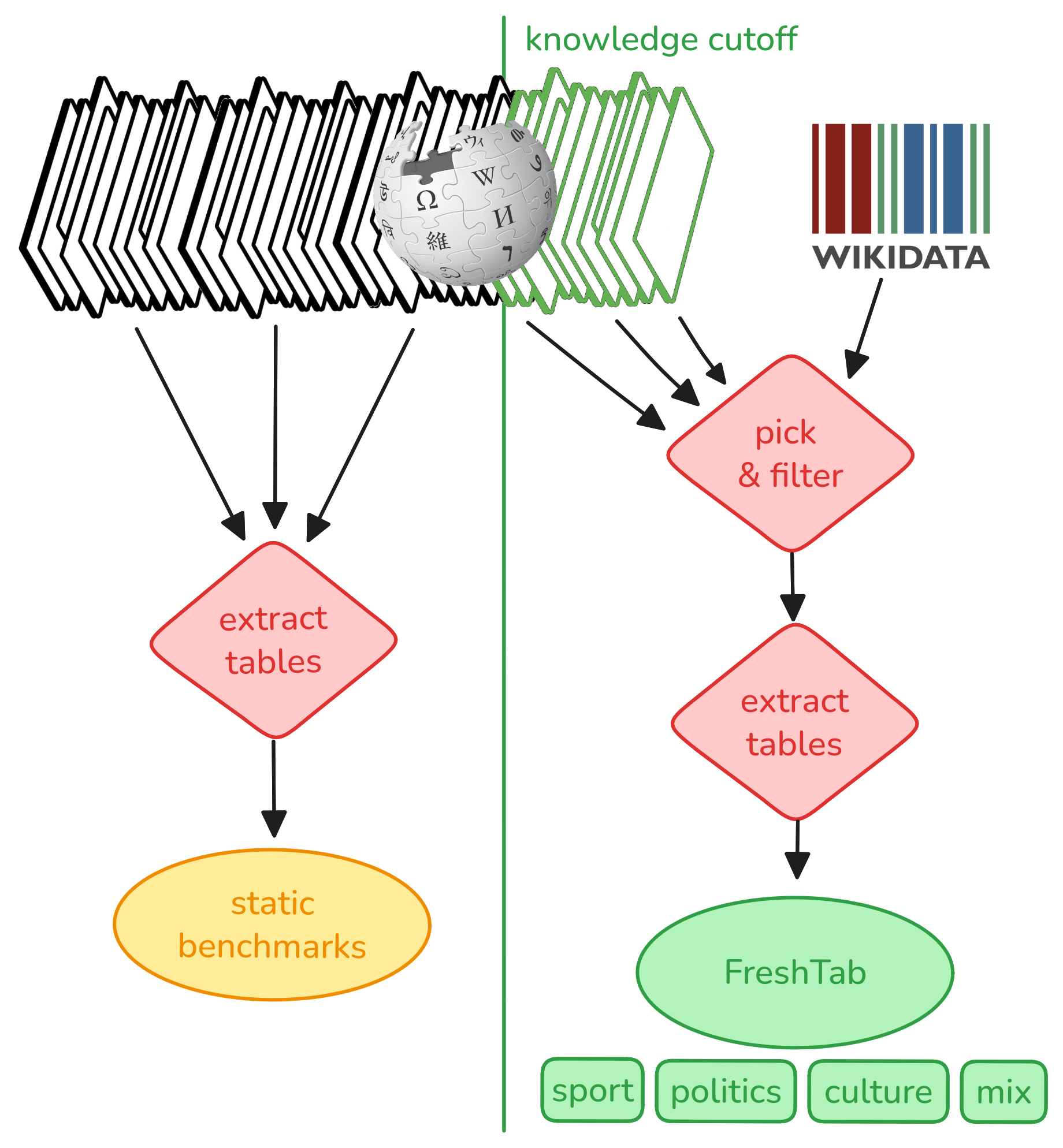}
    \caption{Schema of the FreshTab method}
    \label{fig:FreshTab}
\end{figure}

We directly address these problems and present \emph{FreshTab}, an approach for obtaining up-to-date benchmarks for insight generation, 
following prior work on dynamic dataset construction \cite{kasner_beyond_2024, white2024livebench}. 
This dataset family, based on Wikipedia tables, is not affected by the problems of LLM memorization and benchmark contamination, as the underlying tables are newer than the LLM's knowledge cutoff date, see Figure \ref{fig:FreshTab}. 
We introduce basic domain labels for each table, allowing for domain-specific evaluation insights. The datasets can be generated in any Wikipedia language and configured along multiple parameters.

Our main contributions are as follows: 
\begin{itemize}[left=0pt,itemsep=0.3em,topsep=0.3em,parsep=0pt]
\item We develop \emph{FreshTab} -- a method for creating new table-to-text benchmark datasets based on recent Wikidata/Wikpedia entries, to avoid LLM memorization.
The approach works for any language where a sufficient amount of fresh data is available.

\item We include domain information in the process, to allow for domain-specific evaluation.

\item In experiments using February-May 2025 tables collected with FreshTab, we show that
recent LLMs perform worse than on comparable tables from the earlier LoTNLG/LogicNLG benchmark \citep{zhao-etal-2023-investigating,chen-etal-2020-logical} based on automatic metrics. However, this effect is less pronounced in LLM evaluation and absent in human evaluation, indicating a potential metric bias. We show that domain-balanced data are more challenging than the sport-heavy data used by the previous benchmarks. A LLM evaluation of insights for Russian, German and French tables shows similar performance to English.
\end{itemize}

FreshTab is publicly available and automatically collects a new 
dataset version each month.\footnote{\url{https://github.com/Kristyna-Navitas/FreshTab}}

\section{Related work}

\paragraph{Insight generation}
Approaches for generating insights from tables have been developed alongside other data-to-text NLG systems for decades \cite{barzilay-lapata-2005-collective}.
The emergence of neural models brought a~lot of research into the area, focusing on end-to-end architectures \cite{wiseman-etal-2017-challenges} that incorporate table-aware training \cite{liu_table--text_2018,xing-wan-2021-structure}, use pretrained LMs \cite{kantharaj-etal-2022-chart}, or both \cite{chen-etal-2020-logical,andrejczuk-etal-2022-table}. 
Most recent approaches to table-to-text use LLMs. 
While \citet{bian_helm_2024} and \citet{li_table-gpt_2023} still focus on finetuning LLMs on tabular tasks, 
 \citet{zhao-etal-2023-investigating} and \citet{perez-etal-2025-llm} successfully apply chain-of-thought prompting without the need for task-specific training. 
However, all previous table-to-text approaches focus on fixed benchmarks, making them susceptible to training data contamination \cite{jacovi-etal-2023-stop,li2024task,oren_proving_2024}.

\paragraph{Dynamic benchmarks}
To counteract the issues of LLM training data contamination, \citet{axelsson-skantze-2023-using} propose modifying benchmarks using counter-factual or fictional entities. This partially solves the issue, but the resulting synthetic data are not realistic, and a potential for a repeated leakage remains (hence the non-public release of GEM 2024 test data; \citealp{mille-etal-2024-2024}).
To remove this limitation, dynamic benchmarks emerged recently: \citet{white2024livebench}'s LiveBench represents a set of general questions or problems for LLMs to solve, updated regularly in a manual fashion. \citet{kasner_beyond_2024} focus specifically on the data-to-text generation task, using open APIs to automatically gather fresh input data in several domains.
Our work extends these approaches for the table insight generation task using automatic selection of recent tables from Wikipedia. Furthermore, it adds domain-sensitive evaluation, following \cite{zhu_domaineval_2025}.


\section{Methodology}
\label{sec:dataset}

\subsection{Benchmark Format}

Unlike previous benchmarks using Wikipedia tables \cite{chen-etal-2020-logical,zhao-etal-2023-investigating}, our benchmark only includes Wikipedia data tables with no human reference texts as obtaining references on-the-fly is not feasible.
Instead, we use reference-free evaluation metrics and human evaluation, following \citet{kasner_beyond_2024}.

In addition to the tables themselves, we include \emph{domain labels}, indicating a broad thematic area (\emph{sport}, \emph{politics}, \emph{culture} or \emph{other}) for each table. 
Following the \textit{LoTNLG} benchmark, we also include a~set of five \emph{logical operation labels} (a subset of \textit{aggregation, all, comparative, count, negation, ordinal, simple, superlative, unique}, see Appendix~\ref{app:list}), to provide a suggestion for the model on the type of insight to generate.\footnote{Unlike in \textit{LoTNLG} where they were based on references, the logical labels are sampled randomly in \emph{FreshTab}.}

\subsection{Benchmark Production Process}

Wikipedia has about 64 million pages,\footnote{\url{https://en.wikipedia.org/wiki/Wikipedia:Size_of_Wikipedia}} making it non-trivial to identify pages which contain tables added after a specific date. Therefore, we identify a~relevant subset of pages heuristically.
Our approach proceeds in the following steps:
\begin{enumerate}[left=0pt,itemsep=0.3em,topsep=0.3em,parsep=0pt]
    \item \label{li:query} We query Wikidata using SPARQL queries with a~handpicked set of concepts and categories, to obtain a list of Wikipedia pages appropriate for scraping. 
    This is done with two distinct multi-step approaches. 
    We follow two strategies for determining if a page is truly new, checking for: (1) pages on
    events taking place between the cutoff date and the present and (2)
    pages that were newly created after a~cutoff date.
    \item \label{li:clean} We scrape these pages for tables, clean them and pick one table per page, based on a pre-set targets on table size in terms of number of rows and columns, as well as non-empty cells.
    \item We filter the resulting pages based on configurable domain balance. Each table is also assigned five random logical operations.
\end{enumerate}

The benchmark generation is fully configurable via YAML;
more details on the individual steps are included in Appendix~\ref{app:data-details}.

\section{Experimental Setup}
\label{sec:experiments}

\subsection{Benchmark Comparison}

To evaluate the usefulness of our method, we compare it to the previous \textit{LoTNLG} benchmark \citep{zhao-etal-2023-investigating}, a subset of the commonly used \emph{LogicNLG} data \cite{chen-etal-2020-logical}, which was available to all LLMs at training time, and is paired with reference insights.
Using \textit{FreshTab}, we created several new benchmarks:
\begin{itemize}[left=0pt,itemsep=0.3em,topsep=0.3em,parsep=0pt]
    \item \textit{FreshTab.2-5/25.en.lot} from February-May 2025, after the knowledge cutoff dates for the most recent LLMs. It has 100 English tables with the same domain distribution as the \textit{LoTNLG} benchmark (73 sport, 13 other, 11 culture, and 3 politics tables), to compare the effect of using new data.
    \item \textit{FreshTab.2-5/25.en.diverse} contains 200 English tables, evenly distributed across the four domains, to evaluate domain-specific performance.
    \item \textit{FreshTab.2-5/25} variations in six other languages with the most articles on Wikipedia,\footnote{\url{https://meta.wikimedia.org/wiki/List_of_Wikipedias}} to assess feasibility of producing non-English datasets.
\end{itemize}
We set the table size limit to approx.~3k characters, so that all tables comfortably fit into LLMs' context sizes. The table parameters were taken from the \textit{LogicNLG} \cite{chen-etal-2020-logical} benchmark tables.

\subsection{Models Evaluated}

We evaluate a broad range of open models for insight generation on both \emph{LoTNLG} and our English \emph{FreshTab.2-5/25.en.\{lot/diverse\}} data:
\textit{Llama~3.3~70B} \cite{grattafiori2024llama3herdmodels}, \textit{Qwen~2.5~72B} \cite{qwen2025qwen25technicalreport}, \textit{Mistral Small~3 24B}\footnote{\url{https://mistral.ai/news/mistral-small-3}}, \textit{Gemma~3~27B} \cite{gemmateam2025gemma3technicalreport}, and reasoning models \textit{Magistral} \cite{rastogi2025magistral}
and \textit{DeepSeek R1 Distill Llama 70B} \cite{deepseekai2025deepseekr1incentivizingreasoningcapability}
All generations use a temperature of 0.7, in line with \citet{zhao-etal-2023-investigating}.
We use all models through \textit{Ollama}\footnote{\url{https://ollama.com/}} with 8-bit quantization, to balance our hardware constraints and performance losses due to quantization \cite{marchisio2024does}. We use structured outputs, i.e., constrain the LLM generation to a predefined schema.\footnote{\url{https://ollama.com/blog/structured-outputs}}

\subsection{Prompting setups}

Following \textit{LoTNLG}~\cite{zhao-etal-2023-investigating}, we run two LLM chain-of-thought prompting setups: 
\begin{itemize}[left=0pt,itemsep=0.3em,topsep=0.3em,parsep=0pt]
    \item \textbf{Direct CoT}. The LLM is given the table and description of one logical operation and asked to generate one insight. This runs five times per table for five logical operations. 
    \item \textbf{Choice}. The LLM is given the table and descriptions of all nine logical operations and asked to generate five insights in one go, selecting operations as needed. 
\end{itemize}

\subsection{Human Evaluation}
\label{sec:human-eval}

We run a crowdsourced human evaluation on a~sample of our data (50 tables from each benchmark) with outputs from four LLMs: Llama, DeepSeek, Gemma and Qwen.
We recruit annotators on the Prolific platform.\footnote{\url{https://app.prolific.co/}}

We ask the annotators to spot and highlight accuracy errors in the insights on the word level, following \citet{kasner_beyond_2024}'s setup. 
We operate with four error categories: \emph{incorrect}, 
\emph{not checkable}, 
\emph{misleading}, 
and \emph{other}. 
Details of error categories are explained in the annotation interface, shown in Appendix~\ref{app:annotation}.

\subsection{Automatic Evaluation}
\label{sec:autom-eval}

We use the standard reference-free automatic metrics for the \textit{LogicNLG} benchmark \citep{liu-etal-2022-plog,zhao-etal-2023-investigating} -- trained table entailment metrics \textit{TAPAS} \citep{herzig-etal-2020-tapas} and \textit{TAPEX} \citep{liu2022tapex}. 
We focus on TAPEX in the paper, as we consider output correctness crucial, and TAPEX is the more reliable of the two.
TAPAS as well as scores for other generation aspects are given in Appendix~\ref{app:results} (self-BLEU \cite{zhu2018texygen}, unique tokens \cite{li-etal-2016-diversity} and Shannon entropy \cite{van-miltenburg-etal-2018-measuring} to measure diversity, percentage of failures, and the average output lengths).

In addition, we ran an LLM-as-a-judge evaluation \citep{gu2024survey} with the Llama 3.3 70B model. 
We crafted the prompt to be as close as possible to the annotation instructions for human evaluators (see Section~\ref{sec:human-eval}).


\section{Results}
\label{sec:results}

\begin{figure}[t]
    \centering
    \includegraphics[width=0.48\textwidth, trim={3mm 0 3mm 0}, clip]{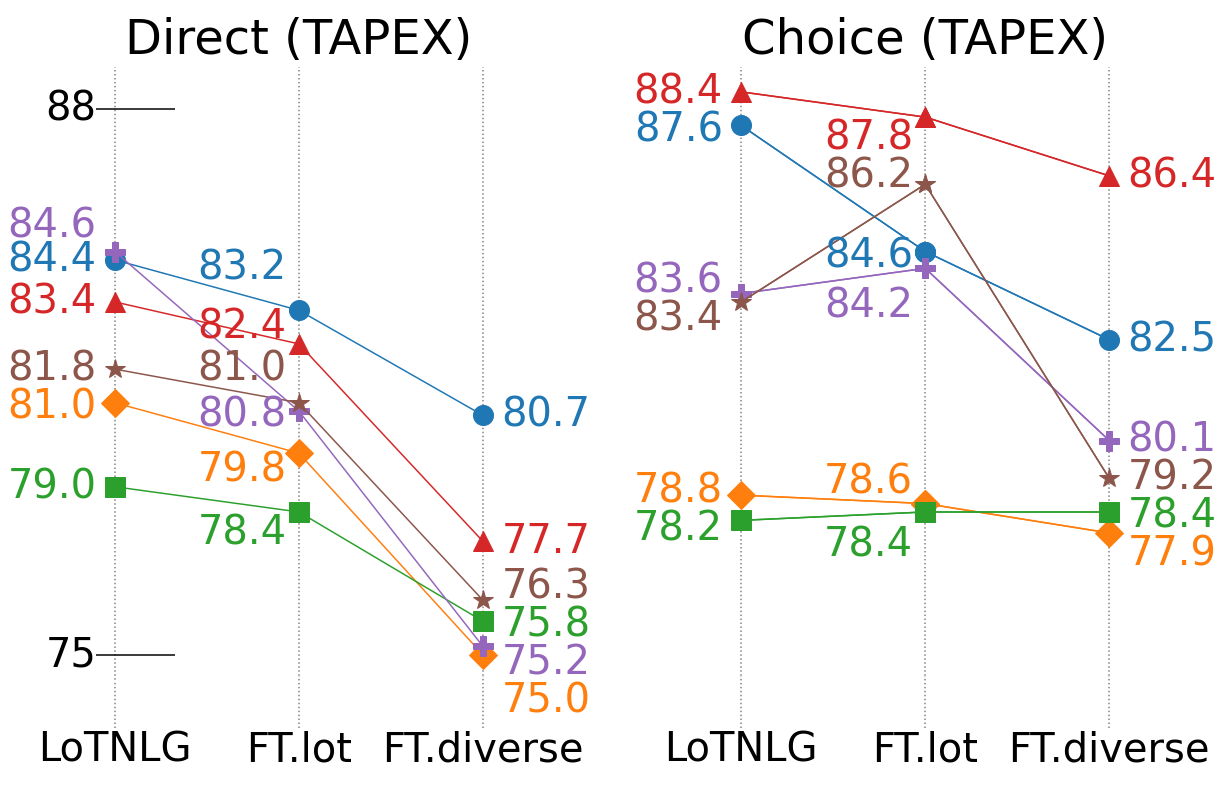}
    \caption{\textit{TAPEX} on \textit{LoTNLG} vs.~\textit{FreshTab.2-5/25.en.lot}vs.~\textit{FreshTab.2-5/25.en.diverse}}
    \label{fig:graph_slope}
\end{figure}

\subsection{TAPEX Performance}
\label{sec:benchmarks}

Based on TAPEX scores in Figure~\ref{fig:graph_slope}, our \textit{FreshTab} benchmark shows more challenging than \textit{LoTNLG} for both prompting setups and most models, especially in the \textit{diverse} domain distribution. 
The \emph{diverse} data proves particularly hard for the DeepSeek and Magistral reasoning LLMs, where the chain-of-thought runs into a dead end and does not produce a valid output in 5\%-10\% cases.

For \emph{Direct CoT}, the performance drop on \emph{FreshTab} is statistically significant for most examined LLMs ($p\leq 0.05$, Z-test for proportions, see  Table~\ref{tab:p_values} in Appendix~\ref{app:results}), with the domain change (\emph{lot} vs. \emph{diverse}) having a stronger effect than the freshness of the tables.

The \textit{Choice} experiment consistently outperforms \textit{Direct CoT}, showing that giving the model more freedom in choosing logical operations pays off.
Performance drop on new data is statistically significant for Llama and Magistral.

\subsection{LLM-as-a-judge Evaluation}

\begin{figure}[t]
    \centering
    \includegraphics[width=0.48\textwidth, trim={3mm 0 3mm 0}, clip]{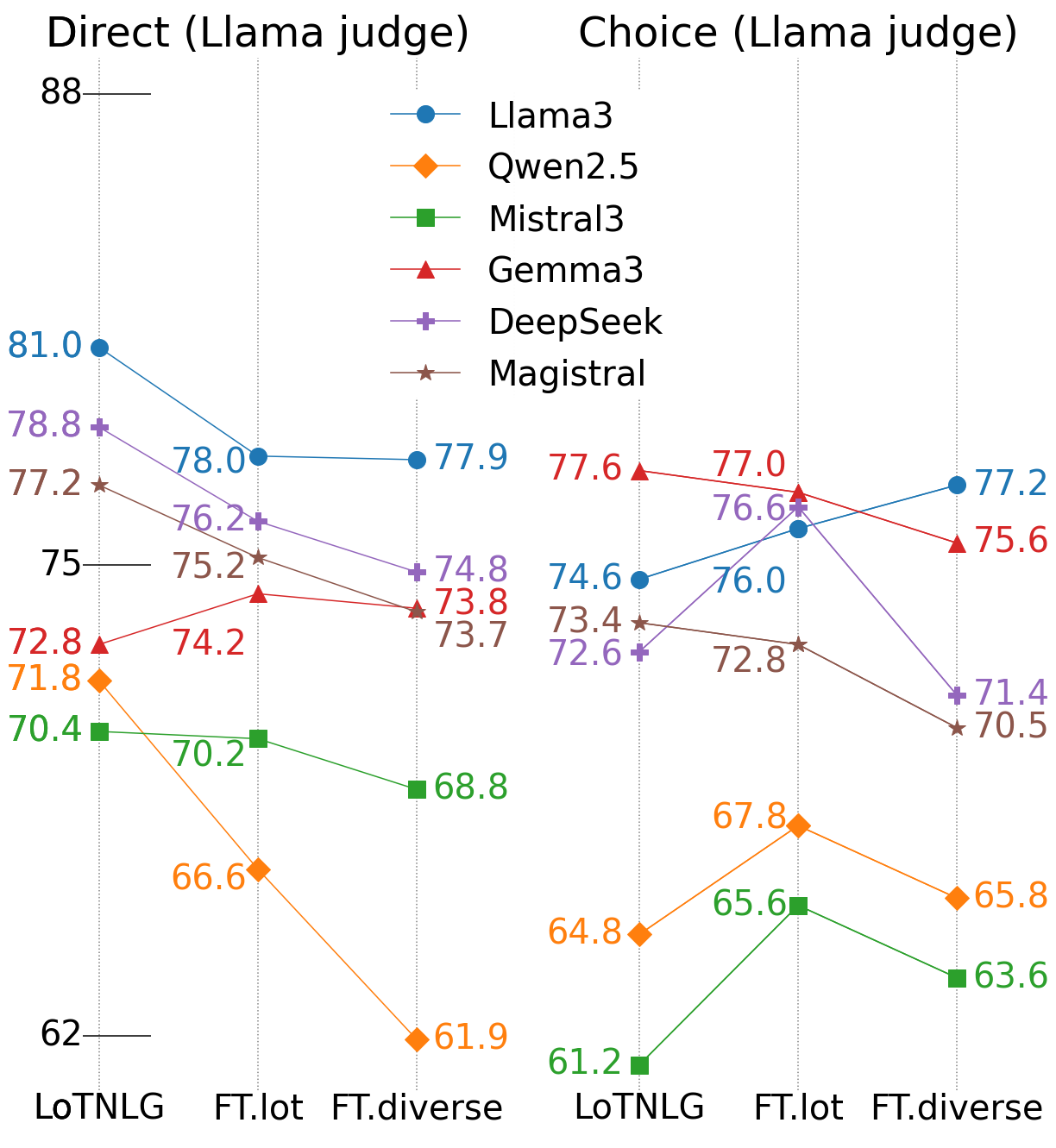}
    \caption{\textit{Llama-as-a-judge} on \textit{LoTNLG} vs.~\textit{FreshTab.2-5/25.en.lot}  vs.~\textit{FreshTab.2-5/25.en.diverse}}
    \label{fig:graph_slope_llm}
\end{figure}

Based on the LLM-judge evaluation in Figure~\ref{fig:graph_slope_llm}, the performance drop on new data is not as straightforward. The scores are lower overall and more varied; 
few differences are statistically significant (Gemma for \emph{Direct}, DeepSeek for \emph{Choice}).
In \emph{Direct}, we often see a performance increase on \emph{FreshTab.2-5/25.en.lot} but a subsequent drop on the \emph{diverse} set. We attribute this to the domain balance.

The scores for \textit{Choice} and \textit{Direct} are mostly similar.
Differences are probably influenced by logical operation choice -- operations picked by LLMs in \emph{Choice} are often different from the ones pre-picked by humans in \emph{Direct} (cf.~Figure~\ref{fig:label_counts} in the Appendix). Overall, all LLMs except Qwen tend to produce \emph{simple} insights more frequently, and Gemma is the most extreme in this regard, gaining higher scores overall.

\begin{figure}[t]
    \centering
    \includegraphics[width=0.4\textwidth]{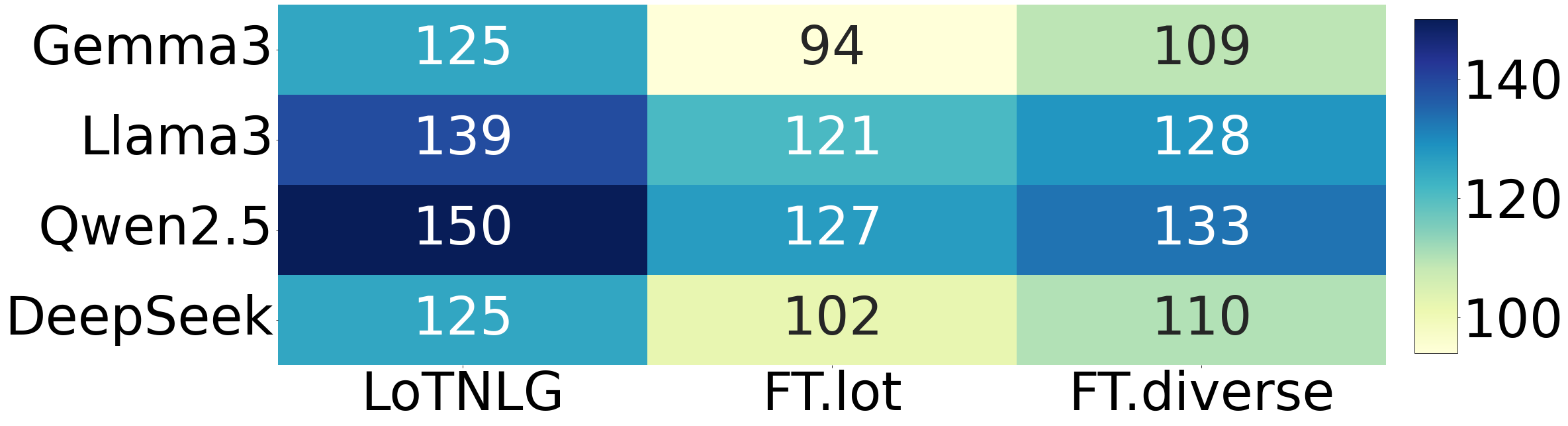}
    \caption{Total number of errors found in human evaluation by model and benchmark 
}
    \label{fig:human_heatmap}
\end{figure}

\begin{figure*}[t]
    \centering
    \includegraphics[width=\textwidth]{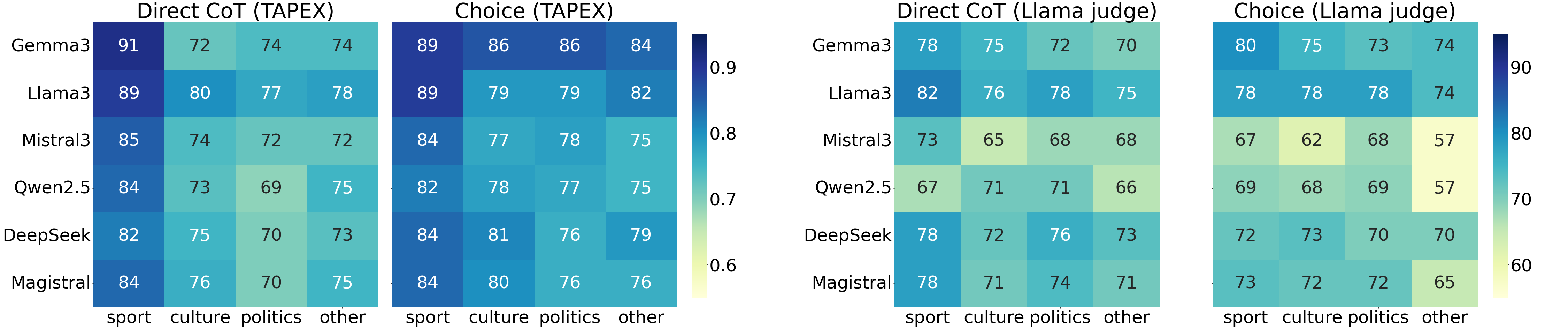}
    \caption{\textit{TAPEX} and \textit{Llama as a judge} on \textit{FreshTab.2-5/25.en.diverse} by domain.}
    \label{fig:graph_heat_tapex}
\end{figure*}

\subsection{Human Evaluation}
\label{sec:human}

Figure~\ref{fig:human_heatmap} shows an overview of our human annotation results (see Table~\ref{tab:prevalences_human} in Appendix~\ref{app:results} for details).
They align better with LLM evaluation than with TAPEX/TAPAS and show an even more consistent trend -- 
the number of errors does not increase on the new data; on the contrary, \emph{FreshTab.2-5/25.en.lot} shows fewer errors overall; the effect is similar in all evaluated LLMs.
The drop on the \emph{diverse} set of \emph{FreshTab} compared to the \emph{lot} set is also clearly visible.

The evaluation differences directly translate to correlations: TAPAS and TAPEX show only low Pearson correlation with humans (0.12 and 0.11).
Based on manual inspection, TAPEX performs better on simple logical operations than on more complex ones.
The LLM-as-a-judge with Llama 3.3 70B produces a~moderate correlation of 0.53 across models and datasets. We compared all other LLMs in the judge setting on the LotNLG set (see Table \ref{tab:Pearson_correlation} in Appendix \ref{app:results}); Llama shows as second highest-correlating but without self-bias.

When we analyze the outputs more closely, we can see that the lower number of errors on \emph{FreshTab} is partly due to logical operation choice. On \emph{LotNLG}, models produce more complex insights (e.g. ``3 episodes have ratings above 16\%.'') by using seen patterns. On \emph{FreshTab} data, they play it safer and produce simpler insights (e.g. ``France is in qualifying group D.''), leading to fewer \emph{aggregation}/\emph{superlative} insights and thus fewer errors.
Errors on \emph{LoTNLG} often concern exact values. With \emph{FreshTab}, models also misinterpret tables (e.g., “Bird [won the most awards] among all the films at Sudbury Film Festival” while the table  only lists awards for the Bird movie), column labels, subtables, row/column switches, or unusual formats (e.g. speech transcript). Numerical operations tend to be less accurate. Reasoning models produce empty outputs more frequently. The models also do not shy away from inconsistent claims, e.g., \textit{"Myanmar has the second-highest number of missing persons, equal to the total across all countries affected by the earthquake"}.

\begin{table}[t]
    \centering\small
    \setlength{\tabcolsep}{5pt}
    \begin{tabular}{lccccccc} 
         \toprule
         language & \cellcolor{blue!25}en & de & fr & sv & nl & ru & es \\
         \midrule
         \cellcolor{blue!25}total &\cellcolor{blue!25}531 &\cellcolor{blue!25} 187&\cellcolor{blue!25} 177&\cellcolor{blue!25} 54&\cellcolor{blue!25} 144&\cellcolor{blue!25} 106&\cellcolor{blue!25}159 \\
         sport &\cellcolor{blue!25} 204& 86 & 73 & 21& 94& 34& 35\\
         politics &\cellcolor{blue!25} 142& 27 & 61 & 1& 4& 9& 19\\
         culture &\cellcolor{blue!25} 109& 51 & 25 & 3& 13& 29& 87\\
         other &\cellcolor{blue!25} 48& 23 & 18 & 29 & 33& 34& 18\\
         \bottomrule
    \end{tabular}
    \caption{Count of pages with new tables for language variations of \textit{FreshTab.2-5/25} (4 months period).}
    \label{tab:languages}
\end{table}

\subsection{Comparison of Domains}
\label{sec:domains}

Figure~\ref{fig:graph_heat_tapex} shows that TAPEX and LLM-judge performance varies across domains. 
With TAPEX, the difference between the \textit{sport} domain and lowest performing domain is statistically significant for all models in the \textit{Direct CoT} experiment and for all except Gemma and Qwen in \textit{Choice} ($p\leq 0.05, $ Z-test for proportions).
For LLM-judge, the differences are only significant for Mistral and Qwen.

With models constrained by pre-set random logical operations in \emph{Direct CoT}, we see \emph{sport} performing mostly better than other domains. 
For \textit{Choice}, TAPEX gets more even across domains as models can pick logical operations. 
LLM-judge reveals that only some models use the larger freedom favorably, with mixed gains and losses.

\subsection{Other Languages}
\label{sec:languages}

Table~\ref{tab:languages} demonstrates that usably-sized datasets, albeit smaller than English, can be produced in other popular Wikipedia languages using \emph{FreshTab}.
We generated insights for three other diverse but high-resource languages from \textit{FreshTab.2-5/25.(de/fr/ru).diverse} and evaluated them with LLM-as-a-judge, as TAPAS and TAPEX cannot be used directly.
The scores are mostly consistent across models; slightly lower for German, similar to English for Russian and slightly higher for French. However, this very much depends on the composition of the new data.

Full results are in Table~\ref{tab:selected_languages} in Appendix~\ref{app:results}.

\section{Conclusion}

We present \emph{FreshTab}, a method for producing live benchmark datasets for table insight generation from Wikipedia, 
enabling easy evaluation of LLMs on unseen data and supporting domain balance and non-English languages.
Our experiments confirm that LLMs behave differently on the new data.
We also found poor performance of automatic metrics, with LLM-judges showing more reliable.


\section*{Limitations}
\label{sec:limitations}
    We use fairly standard generation LLM parameters shared across all steps and consider our setup to be a reasonable baseline.
    We adopted the labels for our general ideas from~\cite{chen-etal-2020-logical,zhao-etal-2023-investigating} but the logical operation categorization is not complete or optimal.
    However, using nine diverse logical operations allowed us to have some degree of controllability and a known source of diversity.
    We acknowledge that the current prompting strategy could be refined and optimized, which we consider as future work. 
%
Some of the data novelty effect may have been compromised by new articles being only translated from another language.
This was only discovered for a~single example, but needs to be further evaluated.

\subsection*{Ethics Statement}
\label{sec:ethics}
The human annotation experiment was approved by an internal ethics committee. The annotators were awarded £9 for the annotation task 
which was estimated to take 60 minutes, in line with the Prolific platform's recommendation.

\section*{Acknowledgments}

This research was co-funded by the European Union (ERC, NG-NLG, 101039303), the National Recovery Plan funded project MPO 60273/24/21300/21000 CEDMO 2.0 NPO, and Charles University project SVV 260 698. It used resources provided by the LINDAT/CLARIAH-CZ Research Infrastructure (Czech Ministry of Education, Youth, and Sports project No.~LM2023062).

\bibliography{custom,anthology}

\clearpage
\appendix

\section{Data collection details}
\label{app:data-details}

Further details for the data collection steps:

\paragraph{Data choice (Step~\ref{li:query}).} We choose the set of concepts and categories by exploration to cover the types of pages that tend to include tables. The tables are picked so that their contents could not have been known before the cutoff date since the page was either non-existent then, or it covers an event (e.g., election, sports competition, book release) that only took place after the cutoff date, and thus its specifics could not have been known before. We check the wikipage's first creation date, to avoid updated entities.
We also abstain from getting largely empty tables relating to future events.

\paragraph{Table selection (Step~\ref{li:clean}).} The table selection includes removing noisy, small, and mostly empty tables based on configurable thresholds. The cleaning step shortens very long tables, simplifies multicolumn names, removes references, consolidates non-values, removes unreasonably long text entries, and empty columns and rows.

\section{Full results}
\label{app:results}

\begin{figure}[t]
    \centering
    \includegraphics[width=0.46\textwidth]{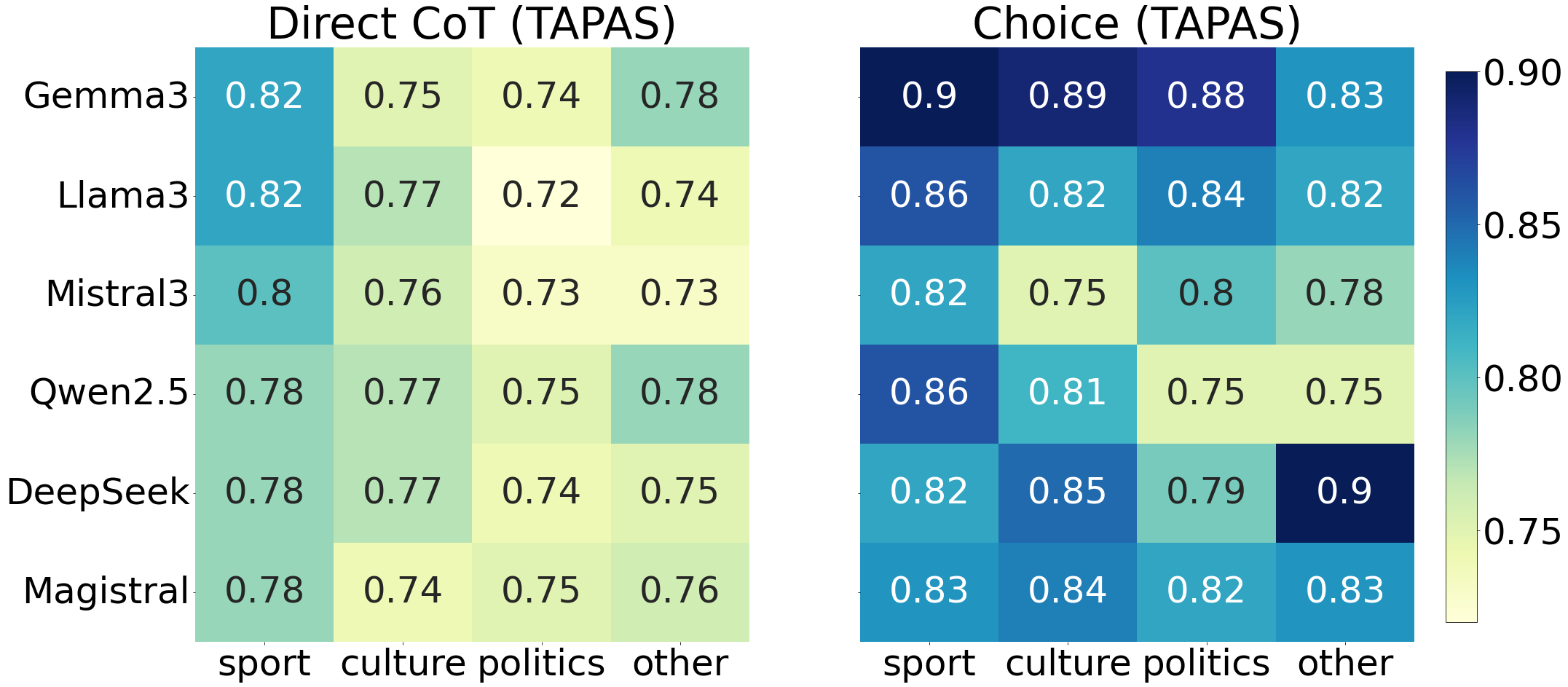}
    \caption{\textit{TAPAS} by domain on \textit{FreshTab.2-5/25.en.diverse}}
    \label{fig:graph_heat_tapas}
\end{figure}

The following tables show our experiments in full: Table~\ref{tab:direct} for the \emph{Direct CoT} experiment and Table~\ref{tab:choice} for the \emph{Choice} experiment. 
P-values for the Z-test for proportions \cite{Walpole2010-mk} on the TAPEX metric between the individual benchmarks are given in Table~\ref{tab:p_values}. Table \ref{tab:operations_acc} shows the TAPEX metric separately for each logical operation.
In addition to the TAPEX metric \cite{liu2022tapex} reported in the main paper, we report TAPAS \cite{herzig-etal-2020-tapas} in Figure~\ref{fig:graph_heat_tapas}.
Note that \textit{TAPEX} treats empty results as not-entailed, as opposed to the \textit{TAPAS} metric that treats these as correct.

To measure insights' diversity, we further report self-BLEU, i.e., BLEU when comparing insights against each other \cite{zhu2018texygen}. Lower self-BLEU means greater diversity.
A further measures of diversity are the average number of unique tokens per insight
and Shannon entropy \cite{van-miltenburg-etal-2018-measuring}.
We also measure the percentage of empty/failed outputs and the average length of the produced insights (in characters).

The full results of human annotation are in Table \ref{tab:factuality_human} showing the percentage of incorrect and misleading insights together, as annotators sometimes used them interchangeably (with gray experiments having too low count to be statistically significant); and Table \ref{tab:prevalences_human} showing the actual counts for all annotated categories.

\begin{figure*}[t]
    \centering
    \includegraphics[width=1\textwidth]{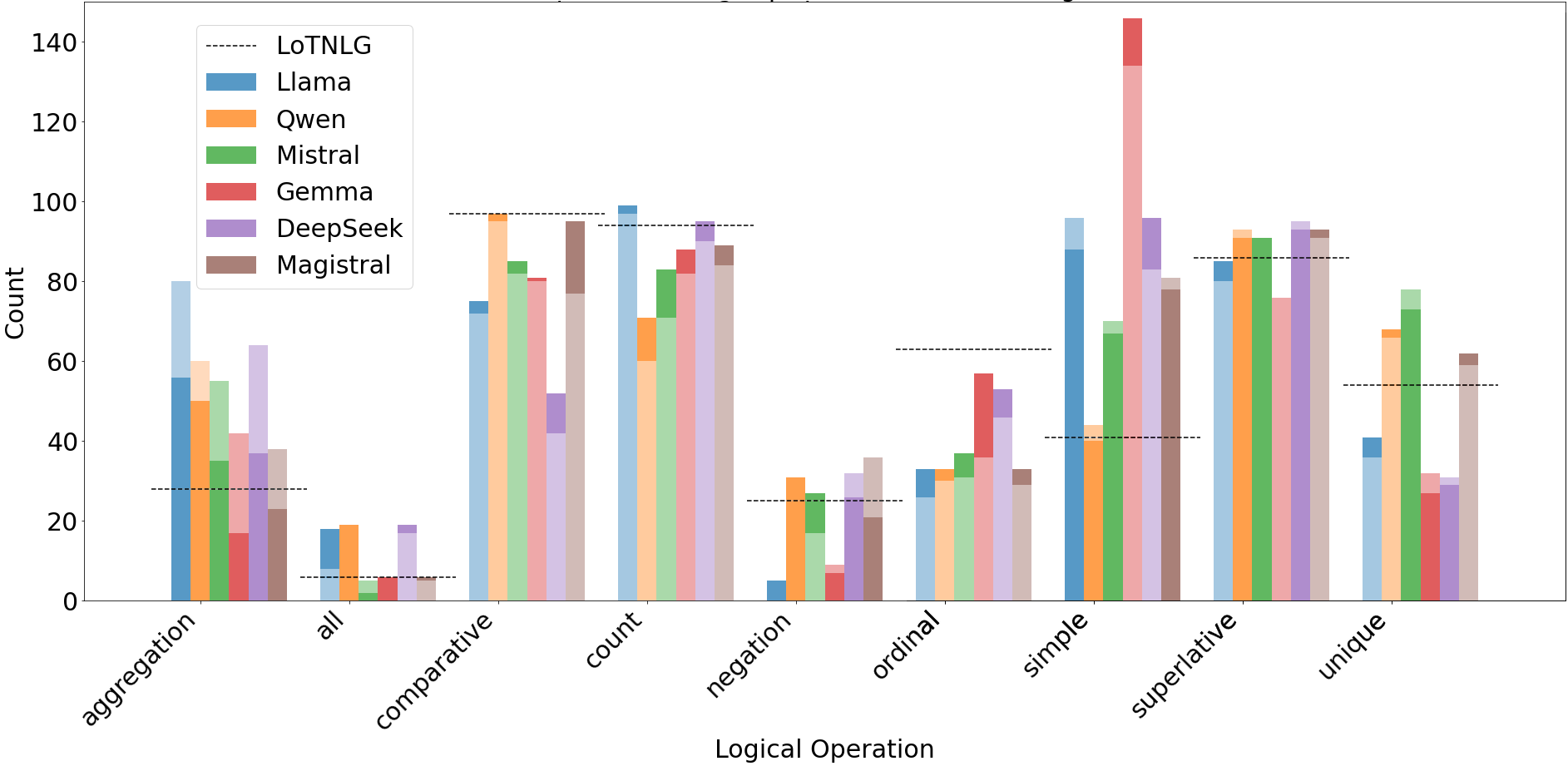}
    \caption{Comparison of logical operation counts. Given from \textit{LoTNLG} (line) and chosen in \textit{Choice} experiment for \textit{LoTNLG} (light) and \textit{FreshTab} (saturated)}
    \label{fig:label_counts}
\end{figure*}

Figure \ref{fig:label_counts} depicts the counts for specific logical operations picked by the different models in the \textit{Choice} experiment related to the human-picked operations in the LoTNLG dataset (horizontal line).

The complete results for all languages tested are in Table \ref{tab:selected_languages}. 
Pearson correlations of different LLM judges with human annotations are in Table \ref{tab:Pearson_correlation}, where the reasoning models were tested only on one set of data due to the high number of tokens generated and not showing a better correlation for it. 

\begin{table*} []
    \centering\small
    \begin{tabular}{lcccccccc}
    \toprule
         model & empty & TAPAS & TAPEX & self-BLEU4 & unique tokens & avg len & entropy\\
         \midrule 
         \multicolumn{8}{c}{\textit{LoTNLG} benchmark}\\
         Gemma 3 & 0.00 & 89.8 & 83.4 & 0.64 & 36 & 86 & 4.71 \\
         Llama 3.3 & 0.00 & 87.0 & 84.4 & 0.56 & 42 & 95 & 5.23 \\
         Mistral & 0.00 & 75.4 & 79.0 & 0.28 & 48 & 88 & 5.11 \\
         Qwen 2.5 & 0.01 & 86.0 & 81.0 & 0.54 & 45 & 102 & 5.18 \\  
         DeepSeek & 0.01 & 85.8 & 84.6 & 0.50 & 39 & 85 & 4.76\\  
         Magistral & 0.02 & 82.4 & 81.8 & 0.41 & 42 & 81 & 4.71 \\  
         \midrule
         \multicolumn{8}{c}{\textit{FreshTab.2-5/25.en} benchmark}\\
         Gemma 3 & 0.00 & 77.4 & 82.4 & 0.39 & 46 & 98 & 5.38 \\
         Llama 3.3 & 0.00 & 76.6 & 83.2 & 0.34 & 49 & 99 & 5.31  \\
         Mistral & 0.00 & 81.4 & 78.4 & 0.13 & 49 & 77 & 5.33 \\
         Qwen 2.5 & 0.00 & 78.4 & 79.8 & 0.35 & 53 & 111 & 5.62 \\
         DeepSeek & 0.03 & 78.2 & 80.8 & 0.33 & 45 & 89 & 5.33 \\  
         Magistral & 0.01 & 79.4 & 81.0 & 0.27 & 48 & 86 & 5.50 \\ 
         \midrule
         \multicolumn{8}{c}{\textit{FreshTab.2-5/25.en.diverse} benchmark}\\
         Gemma 3 & 0.01 & 77.3 & 77.7 & 0.36 & 47 & 101 & 4.73 \\ 
         Llama 3.3 & 0.00 & 76.3 & 80.7 & 0.30 & 51 & 105 & 5.21 \\
         Mistral & 0.00 & 75.7 & 75.8 & 0.25 & 49 & 90 & 5.13 \\ 
         Qwen 2.5 & 0.01 & 77.2 & 75.0 & 0.30 & 56 & 115 & 5.55 \\
         DeepSeek & 0.07 & 75.9 & 75.2 & 0.29 & 47 & 90 & 4.69 \\  
         Magistral & 0.05 & 75.5 & 76.3 & 0.24 & 48 & 89 & 4.95 \\ 
         \bottomrule
    \end{tabular}
    \caption{Automatic metrics for the \textit{Direct CoT} experiment}
    \label{tab:direct}
\end{table*}

\begin{table*} []
    \centering\small
    \begin{tabular}{lcccccccc}
    \toprule
         model & empty & TAPAS & TAPEX & self-BLEU4 & unique tokens & avg len & entropy\\
         \midrule
         \multicolumn{8}{c}{\textit{LoTNLG} benchmark}\\
         Gemma 3 & 0.00 & 87.2 & 88.4 & 0.15 & 52 & 88 & 5.35 \\
         Llama 3.3 & 0.00 & 88.8 & 87.6 & 0.18 & 61 & 110 & 5.57 \\
         Mistral & 0.00 & 81.8 & 78.2 & 0.13 & 52 & 82 & 5.42 \\
         Qwen 2.5 & 0.00 & 80.0 & 78.8 & 0.17 & 62 & 103 & 5.61 \\
         DeepSeek & 0.01 & 83.2 & 83.0 & 0.14 & 51 & 81 & 5.29\\  
         Magistral & 0.05 & 82.2 & 83.4 & 0.16 & 51 & 79 & 5.44 \\ 
         \midrule
         \multicolumn{8}{c}{\textit{FreshTab.2-5/25.en} benchmark}\\
         Gemma 3 & 0.00 & 87.0 & 87.8 & 0.16 & 50 & 83 & 5.49 \\
         Llama 3.3 & 0.00 & 83.4 & 84.6 & 0.20 & 60 & 109 & 5.89 \\
         Mistral & 0.00 & 81.4 & 78.4 & 0.13 & 49 & 77 & 5.33 \\
         Qwen 2.5 & 0.00 & 82.4 & 78.6 & 0.19 & 60 & 101 & 5.65 \\
         DeepSeek & 0.06 & 83.4 & 84.2 & 0.17 & 47 & 73 & 4.23 \\  
         Magistral & 0.03 & 87.6 & 86.2 & 0.16 & 49 & 78 & 5.72 \\  
         \midrule
         \multicolumn{8}{c}{\textit{FreshTab.2-5/25.en.diverse} benchmark}\\
         Gemma 3 & 0.00 & 87.6 & 86.4 & 0.16 & 53 & 92 & 5.41 \\
         Llama 3.3 & 0.00 & 83.5 & 82.5 & 0.18 & 63 & 115 & 5.70 \\
         Mistral & 0.00 & 78.9 & 78.4 & 0.12 & 53 & 87 & 5.48 \\
         Qwen 2.5 & 0.00 & 79.5 & 77.9 & 0.18 & 61 & 106 & 5.87 \\
         DeepSeek & 0.10 & 81.5 & 80.1 & 0.16 & 49 & 79 & 6.01 \\  
         Magistral & 0.11 & 83.0 & 79.2 & 0.17 & 50 & 80 & 5.35\\  
         \bottomrule
    \end{tabular}
    \caption{Automatic metrics for the \textit{Choice CoT} experiment}
    \label{tab:choice}
\end{table*}

\begin{table*} []
    \centering\small
    \begin{tabular}{lccc}
    \toprule
         model & \textit{LoTNLG} vs \textit{Diverse} & \textit{LoTNLG} vs \textit{FreshTab} & \textit{FreshTab} vs \textit{Diverse}\\
         \hline
         \multicolumn{4}{c}{\textit{Direct} CoT experiment}\\
         Gemma 3 & 0.01 & 0.67 & 0.03 \\
         Llama 3.3 & 0.08 & 0.61 & 0.24 \\
         Mistral & 0.17 & 0.82 & 0.26 \\
         Qwen 2.5 & 0.01 & 0.63 & 0.04 \\
         DeepSeek & 0.00 & 0.11 & 0.02 \\
         Magistral & 0.02 & 0.75 & 0.04 \\
         \midrule
         \multicolumn{4}{c}{\textit{Choice} CoT experiment}\\
         Gemma 3 & 0.28 & 0.77 & 0.45 \\
         Llama 3.3 & 0.01 & 0.17 & 0.31 \\
         Mistral & 0.93 & 0.91 & 1.00 \\
         Qwen 2.5 & 0.69 & 0.94 & 0.76 \\
         DeepSeek & 0.18 & 0.61 & 0.05 \\
         Magistral & 0.05 & 0.22 & 0.00 \\
         \bottomrule
    \end{tabular}
    \caption{Statistical significance between datasets}
    \label{tab:p_values}
\end{table*}

\begin{table*} []
    \centering\small
    \begin{tabular}{lccccccccc}
    \toprule
         model & aggregation	&all &comparative &count &negation & ordinal & simple & superlative & unique\\
         \midrule
         \multicolumn{9}{c}{\textit{LoTNLG} benchmark}\\
         Gemma & 80.0 &77.8 &81.4 &71.6 & 60.7 & 85.9 & 87.8 & 87.1 & 84.5 \\
         Llama & 83.3 &77.8 &79.4 &77.3 & 75.0 & 89.1 & 95.1 & 97.6 & 77.6 \\
         Mistral & 86.7 &88.9 & 88.7 & 84.1 & 71.4 &78.1 &95.1 &88.2 & 75.9 \\
         Qwen 2.5 & 90.0 & 88.9 &82.5 & 86.4 &64.3 &82.8 &85.4 &87.1 &58.6 \\
         DeepSeek & 90.0 & 55.6 &78.4 &88.6 &67.9 &89.1 &92.7 & 95.3 &72.4 \\
         Magistral & 89.7 &77.8 &88.5 &77.9 &60.7 &79.4 &95.1 &91.6 &70.7 \\

         \midrule
         \multicolumn{9}{c}{\textit{FreshTab.2-5/25.en} benchmark}\\
         Gemma 3 & 89.2 & 66.7 & 82.0 &88.7 & 54.2 & 96.8 & 86.2 & 80.9 & 93.1 \\
         Llama 3.3 & 92.3 & 75.0	& 82.0 & 81.1 & 59.3 & 91.9 & 91.4 & 89.4 & 84.5 \\
         Mistral & 90.8 & 43.8 & 76.0 & 86.8 & 55.9 & 83.9 & 94.8 & 89.4 & 84.5 \\
         Qwen 2.5 & 83.1 & 60.4 & 82.0 & 86.8 & 55.9 & 91.9 & 94.8 & 87.2 & 74.1 \\
         DeepSeek & 93.8 & 64.6 & 86.0 & 78.8 & 66.1 & 93.5 & 84.5 & 84.8 &74.1 \\
         Magistral & 87.5 & 72.9 & 84.0 & 73.1 & 61.0 & 93.4 & 91.4 & 87.2 & 81.0 \\
         \midrule
         \multicolumn{9}{c}{\textit{FreshTab.2-5/25.en.diverse} benchmark}\\
         Gemma 3 & 86.4 & 68.3 & 82.2 & 74.8 & 49.6 & 90.4 & 81.9 & 85.1 & 80.9 \\
         Llama 3.3 & 87.3 & 74.0 & 85.0 & 75.7 & 62.2 & 89.6 & 88.8 & 80.2 & 82.6 \\
         Mistral & 84.5 & 54.8 & 78.5 & 77.5 & 47.9 & 78.4 & 92.2 & 84.2 & 85.2 \\
         Qwen 2.5 & 79.1 & 54.8 & 82.2 & 76.7 & 58.0 & 80.0 & 86.2 & 92.1 & 67.0 \\
         DeepSeek & 87.3 & 60.6 & 78.5 & 71.8 & 63.0 & 90.2 & 71.6 & 88.0 & 67.8 \\
         Magistral & 79.4 & 71.2 & 82.1 & 76.0 & 55.5 & 81.3 & 88.7 & 84.2 & 74.6 \\
         \bottomrule
    \end{tabular}
    \caption{TAPEX for logical operations for \textit{Direct} CoT experiment.}
    \label{tab:operations_acc}
\end{table*}

\begin{table*} []
    \centering\small
    \begin{tabular}{lccccc}
    \toprule
          model & sport & culture & politics & other \\
          \midrule
          \multicolumn{5}{c}{\textit{LoTNLG} benchmark}\\
          counts & 190 & \color{gray} 20 & \color{gray} 15 &\color{gray}  25\\
          Gemma 3 & 0.35 &\color{gray}  0.25 & \color{gray} 0.20 & \color{gray} 0.48  \\
          Llama 3.3 & 0.44 & \color{gray} 0.30 &\color{gray}  0.40 &\color{gray}  0.44\\
          Qwen 2.5 & 0.46 &\color{gray}  0.45 & \color{gray} 0.27 & \color{gray} 0.28  \\
          DeepSeek & 0.35 &\color{gray}  0.45 &\color{gray}  0.47 & \color{gray} 0.36  \\ 
          \midrule
          \multicolumn{5}{c}{\textit{FreshTab.2-5/25.en} benchmark}\\
          counts & 160& 50& \color{gray} 5& \color{gray} 35\\
          Gemma 3 & 0.29 & 0.24 & \color{gray} 0.40 & \color{gray} 0.17 \\
          Llama 3.3 & 0.39 & 0.26 &\color{gray}  0.80 & \color{gray} 0.17  \\
          Qwen 2.5 & 0.38 & 0.36 &\color{gray}  0.60 & \color{gray} 0.26 \\
          DeepSeek & 0.24 & 0.28 & \color{gray} 0.40 & \color{gray} 0.11 \\
          \midrule 
          \multicolumn{5}{c}{\textit{FreshTab.2-5/25.en.diverse} benchmark}\\
          counts & 55& 75& 65& 55\\
          Gemma 3 & 0.36 & 0.25 & 0.22 & 0.40 \\
          Llama 3.3 & 0.53 & 0.24 & 0.35 & 0.29 \\
          Qwen 2.5 & 0.49 & 0.25 & 0.37 & 0.45 \\
          DeepSeek & 0.36 & 0.28 & 0.32 & 0.31 \\ 
          \bottomrule
    \end{tabular}
    \caption{Percentage of incorrect+misleading insights from human annotation by domains.}
    \label{tab:factuality_human}
\end{table*}

\begin{table*} []
    \centering\small
    \begin{tabular}{lccccc}
    \toprule
         model & Incorrect & Misleading & Not checkable & Other \\
         \midrule
         \multicolumn{5}{c}{\textit{LoTNLG} benchmark}\\
         Gemma 3 & 67 & 20 & 20 & 18  \\
         Llama 3.3 & 83 & 23 & 13 & 20\\
         Qwen 2.5 & 82 & 26 & 21 & 21  \\
         DeepSeek & 68 & 23 & 15 & 19  \\ 
         \midrule
         \multicolumn{5}{c}{\textit{FreshTab.2-5/25.en} benchmark}\\
         Gemma 3 & 49 & 18 & 10 & 17 \\
         Llama 3.3 & 58 & 28 & 17 & 18  \\
         Qwen 2.5 & 73 & 17 & 23 & 14 \\
         DeepSeek & 46 & 13 & 21 & 22 \\
         \midrule
         \multicolumn{5}{c}{\textit{FreshTab.2-5/25.en.diverse} benchmark}\\
         Gemma 3 & 59 & 16 & 11 & 23 \\
         Llama 3.3 & 64 & 22 & 18 & 24 \\
         Qwen 2.5 & 76 & 19 & 12 & 26 \\
         DeepSeek & 70 & 9 & 14 & 17 \\ 
         \bottomrule
    \end{tabular}
    \caption{Factuality span annotations prevalences from human annotation.}
    \label{tab:prevalences_human}
\end{table*}

\begin{table}[t]
    \centering\small
    \begin{tabular}{lcccc} 
          \toprule
          & en & de & fr & ru \\
          \midrule
         \multicolumn{5}{c}{Choice experiment} \\
         \midrule
         Gemma & 75.6 & 74.4 & 79.6 & 83.2 \\
         Llama & 77.2 & 71.4 & 76.8 & 76.2 \\
         Mistral & 63.6 & 61.4 & 71.2 & 68.2 \\
         Qwen & 65.8 & 66.2 & 69.4 & 67.4 \\
         DeepSeek & 71.5 & 56.0 & 69.6 & 62.0 \\
         Magistral & 70.5 & 72.0 & 69.8 & 64.4\\
         \midrule
         \multicolumn{5}{c}{Direct experiment} \\
         \midrule
         Gemma & 73.8 & 74.0 & 80.8 & 72.8 \\
         Llama & 77.9 & 76.8 & 81.6 & 78.8 \\
         Mistral & 68.8 & 69.0 & 70.4 & 70.0 \\
         Qwen & 69.1 & 70.8 & 74.0 & 72.8 \\
         DeepSeek & 74.8 & 71.8 & 78.8 & 76.0 \\
         Magistral & 73.7 & 70.6 & 75.0 & 75.0 \\
         \bottomrule
    \end{tabular}
    \caption{Factuality of generations for selected languages with Llama-as-a-judge for \textit{FreshTab.2-5/25.diverse}.}
    \label{tab:selected_languages}
\end{table}

\begin{table}[]
    \centering\small
    \begin{tabular}{lcccc} 
    \toprule
         judge / insight & Gemma & Llama & Qwen & DeepSeek \\
         \midrule
         Gemma & 0.82 & 0.52 & 0.61 & 0.43 \\
         Llama & 0.46 & 0.56 & 0.61 & 0.54 \\
         Qwen & 0.47 & 0.40 & 0.42 & 0.47 \\
         Mistral & 0.34 & 0.41 & 0.33 & 0.44 \\
         \midrule
         DeepSeek & - & 0.46 & - & - \\
         Magistral & - & 0.20 & - & - \\
    \bottomrule
    \end{tabular}
    \caption{Pearson correlations between LLM-as-a-judge with different LLMs and human evaluation for LoTNLG dataset.
    }
    \label{tab:Pearson_correlation}
\end{table}

\section{Human Annotation Details}
\label{app:annotation}

The examples for the human annotation are sampled randomly while excluding tables with over 120 characters in the header, to fit into the annotation interface without horizontal scrolling.

We use the \textit{Factgenie} annotation tool \cite{kasner_factgenie_2024}. 
Each annotator is given 3 tables, each paired with 21 insights -- five insights per evaluated model, plus one table-unrelated insight used as an attention check.\footnote{We sample the attention check insights from insights related to different input tables.}

\begin{figure*}[h]
\includegraphics[width=\textwidth]{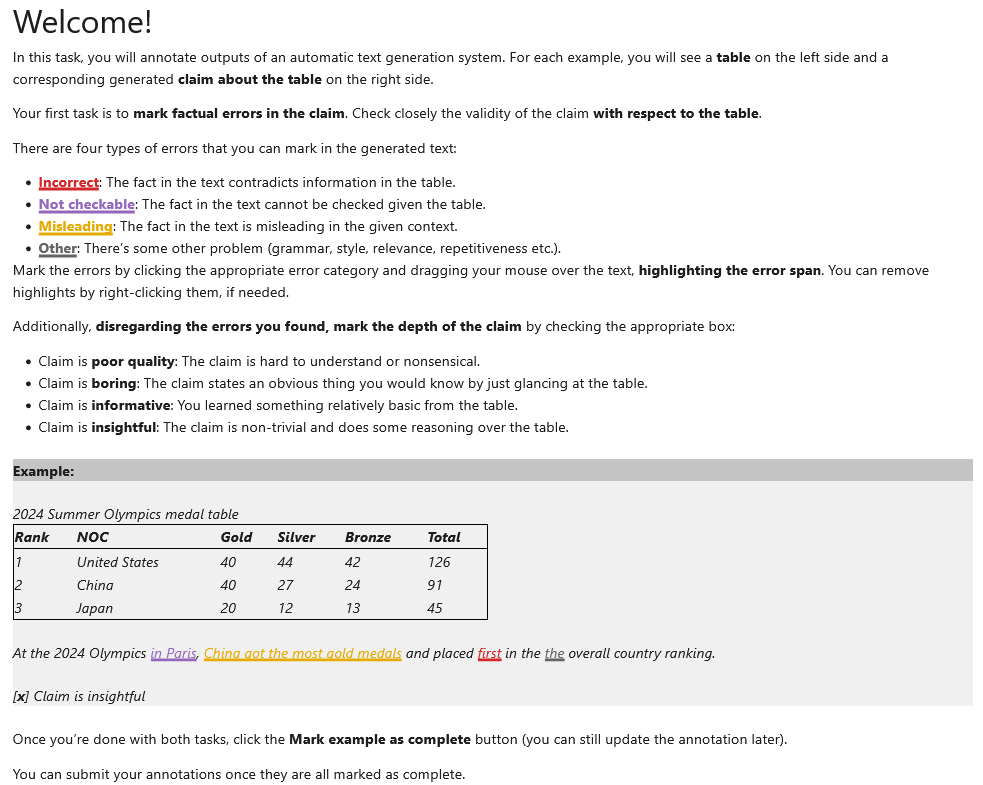}
\caption{Annotation instructions for the human evaluation campaign.}
\label{fig:annot-instruct}
\end{figure*}

\begin{figure*}[h]
\includegraphics[width=\textwidth]{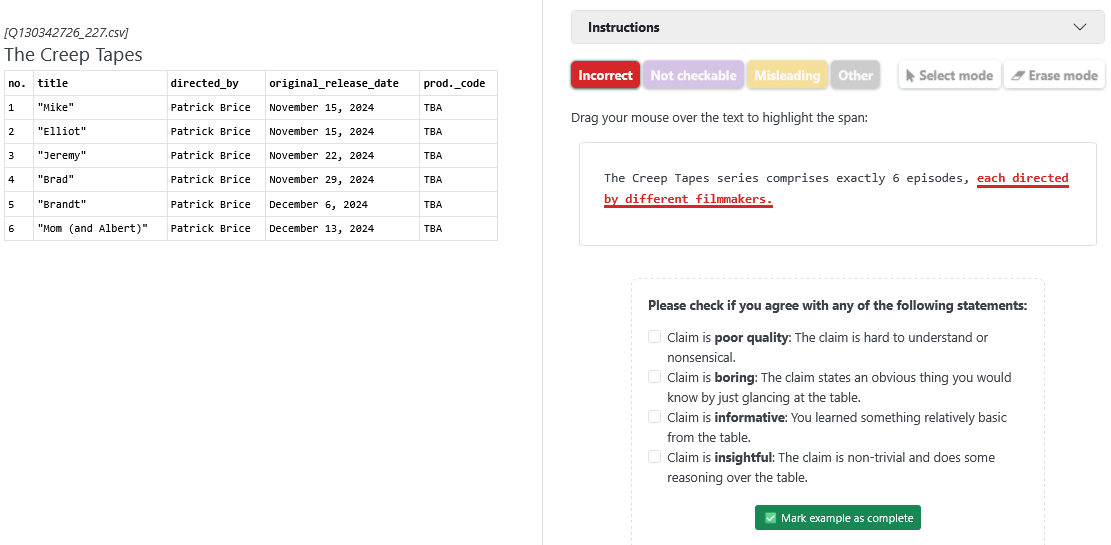}
\caption{Annotation interface, with the table on the left and the annotation form on the right. Annotators can display the instructions by clicking on the top-right collapsible panel.}
\label{fig:annot-interface}
\end{figure*}

Detailed annotation instructions, as shown to the annotators prior to annotation, are given in Figure~\ref{fig:annot-instruct}. The annotation interface is shown in Figure~\ref{fig:annot-interface}.
Annotators were pre-selected based on their country of residence (UK, U.S., Ireland, Australia, New Zaeland), their indicated primary language (English) and good approval rate.
We manually checked whether annotators gave meaningful replies to the attention check instances, and if not, their annotations were replaced by an additionally hired annotator.

\section{Types of logical inferences }
\label{app:list}
We use the following nine logical operations, proposed by \citet{zhao-etal-2023-investigating}:
\begin{itemize}[left=0pt,itemsep=0.3em,topsep=0.3em,parsep=0pt]
\item \emph{aggregation} -- insights that mention aggregate statistics of data such as sums or averages, e.g., average home team score
\item \emph{all} -- insights where all items share a common property, e.g., all games were played on the same date
\item \emph{comparative} -- insights that compare different entities on some property, e.g., comparing the scores of two teams
\item \emph{count} -- knowledge about the number of entities that fulfill some condition, e.g., number of teams that played at a particular venue
\item \emph{negation} -- formulates a negative claim about an entity, e.g., Team A never played against Team B
\item \emph{ordinal} -- indicates the ranking of entities on some aspect, e.g., second largest crowd to watch the match at a venue
\item \emph{simple} -- the sentences which do not involve higher-order operations, e.g.,  Player X is from country Y.
\item \emph{superlative} -- data insights about maximum or minimum values, e.g., highest score by any team
\item \emph{unique} -- insights about distinct values of a column, e.g., the matches were played in different venues
\end{itemize}

\end{document}